\documentclass{nlp-class}

\usepackage{amsmath}
\usepackage{graphicx}

\begin{document}

\setlength{\pdfpageheight}{\paperheight}
\setlength{\pdfpagewidth}{\paperwidth}




\titlebanner{banner above paper title}        
\preprintfooter{short description of paper}   

\title{Transformer based Urdu Handwritten Text Optical Character Reader}

\authorinfo{Mohammad Daniyal Shaiq}
           {FAST NUCES}
           {i191794@nu.edu.pk}
\authorinfo{Musa Dildar Ahmed Cheema}
           {FAST NUCES}
           {i191765@nu.edu.pk}
\authorinfo{Ali Kamal}
           {FAST NUCES}
           {i191865@nu.edu.pk}

\maketitle

\begin{abstract}
Extracting Handwritten text is one of the most important components of digitizing information and making it available for large scale setting. Handwriting Optical Character Reader (OCR) is a research problem in computer vision and natural language processing computing, and a lot of work has been done for English, but unfortunately, very little work has been done for low resourced languages such as Urdu. Urdu language script is very difficult because of its cursive nature and change of shape of characters based on it's relative position, therefore, a need arises to propose a model which can understand complex features and generalize it for every kind of handwriting style. In this work, we propose a transformer based Urdu Handwritten text extraction model. As transformers have been very successful in Natural Language Understanding task, we explore them further to understand complex Urdu Handwriting.
\end{abstract}


\section{Problem statement}
In these current times, there has been an exponential growth in the adoption of technology because of the
promises it makes - easy and fast access to information, accuracy, connectivity and security. Hence, there exists
a need to digitize information in order to preserve, and make better use of it. One of the potential raw information
which needs to be digitized, is handwritten text (HWT), especially low resource text written in low resource
languages such as Urdu. Optical Character Reader (OCR) is one way to digitize printed text, but recognizing
handwritten text is a challenge because of various different writing styles and other irregularities. There has been
a significant amount of work done for English Handwritten Text extraction, however, the same cannot be said
for low resource languages such as Urdu, which in itself has a user base of 200 million people. Additionally, Urdu
has its largest user base in Pakistan, where countless businesses and other processes still involve handwritten
Urdu text. Therefore, there’s a need to build such programs that can digitize Urdu Handwritten text, which can impact the lives of about 200 million people. Urdu, unlike Latin languages, is written in Arabic like script,
hence it is more challenging because of ligatures, joining of letters leading to change of shape, and a vast variety
of writing styles.
\begin{figure}[h!]
  \includegraphics[width=\linewidth]{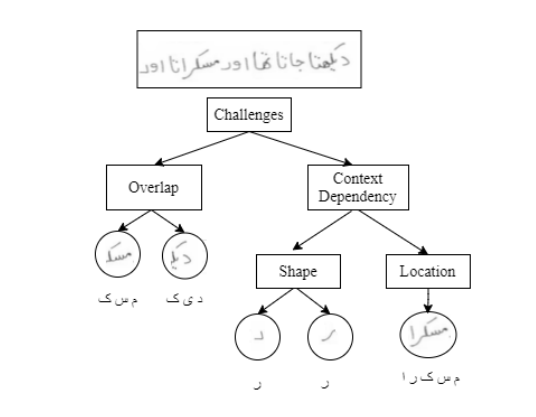}
  \caption{Challenges in extracting Urdu Script, identified by \cite{anjum2020urdu_ohtr}}
\end{figure}
\begin{figure}[h!]
  \includegraphics[width=\linewidth]{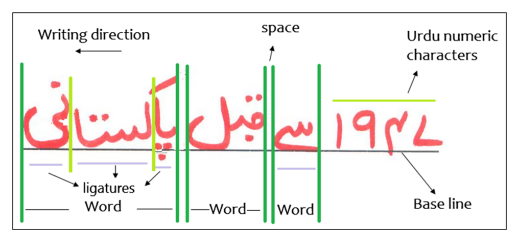}
  \caption{Different characteristics of Urdu Script}
\end{figure}
\section{Introduction}
Urdu language has its second largest user base in the Indian Subcontinent, amounting to about 200 million
people. The Urdu language is written in Arabic like script, Nastaleeq script. Urdu language has about 45
different characters, and these characters are joined while writing which changes their shape. This concept
is known as ligatures. Urdu language has, in total, about 26,000 ligatures \cite{10.1109/ICDAR.2013.212}. Due to this phenomenon,
it makes it exponentially difficult for OCR systems to extract characters from images. Moreover, there are
other complexities in Urdu script, such as lack of word spacing, diagonal writing style, change of shape with
context, and an assortment of different styles. Considering these problems in Urdu script, extracting printed
Urdu text with a standardised style is a challenge for OCR systems, let alone extracting Urdu Handwritten
text. Handwritten Urdu text adds more complexity to the text by adding noise, non standardised format, more
overlap, unequal spacing, inconsistent shapes etc. Arabic language, which has a similar script to Urdu, has
gained a lot of interest from the research community, and in fact there are many commercial products for Arabic
text, but for Urdu there is only one commercially available OCR solution \cite{10.1109/ICDAR.2013.212}, which works only for printed
Urdu text, and so far there has been no commercial product for Urdu Handwriting. Therefore, there exists a
dire need to propose a solution for extracting Urdu Handwriting, which can serve as the pillar for commercial
handwriting extraction tools. However, the problem for Urdu Handwriting extraction is till in it’s early days,
as there are only two publicly available data sets - NUST-UHWD \cite{ul_Sehr_Zia_2021} and UHWD. Written Urdu language is
widely used in the Indian subcontinent, as there’s still a huge gap in digital penetration in the region. Many
forms, cheques, FIR documents, and many historical documents are present in written form, alongside various
business processes. Due to all this, overall process from government processes to private processes are very slow,
as there’s a need of manual verification. Digitizing this problem will have a huge impact in the lives of many
and can help organizations in providing efficient and quick service deliveries.

There are two forms of writing scripts available for languages, cursive and non-cursive. Extraction of non-cursive
handwritten text is easier as compared to cursive as letters are isolated, i-e. they are not joined together, figure
1 illustrates it. Urdu language is written in cursive style, which makes it a difficult problem. There has been
work done for extracting cursive style text such as Arabic, and there are many publicly available data sets for
Arabic language, but they cannot be used for Urdu, because despite Arabic and Urdu being sister languages,
which share an almost similar script, there exists a difference in writing style in both the languages. Urdu is
written in Nastaleeq writing style and Arabic is written in Naskh writing style \cite{malik2005sa}. Nastaleeq writing style is
more complex than Naskh because it has diagonality, filled loops, no fixed baseline, large variation of words/sub
words and false loop. Moreover, the number of ligatures and characters in Urdu language are more than Arabic
language. Therefore, the data set and OCR models for Arabic language cannot certainly satisfy Urdu language.
As of now, collecting a large diverse Urdu Handwritten data set can be a huge contribution
In the handwritten text extraction problem, there are further two divisions, Online Handwriting recognition
and Offline Handwriting recognition. Online Handwriting is the form of writing text on a digital pad using
a stylus, while on the contrary, Offline Handwriting is written on any piece of paper. Recognizing Online
Handwriting is easier, as pixel level information of the handwriting is stored in the computer, which can help in
identification of text, while also removing the noise. On the other hand, Offline Handwriting is written on paper,
which has no other information apart from the image itself, making it much more difficult for the computer to
comprehend. Also, images contain background noise, which adds another layer of difficulty to the problem. We,
in this research, worked on offline handwritten Urdu text, as very little work is done in it. Our
ultimate goal is digitization of Urdu documents, forms, cheques etc. which come under the umbrella of Offline
Urdu Handwriting. Lastly, the most challenging aspect of Urdu Handwriting is change of shape, with change of
position or context. This is illustrated in the figure 2.

\section{Related work}
In \cite{one_direc}, the authors study the problem of Urdu Handwriting detection, and identify the issue faced in extraction
of Urdu handwriting as compared with other languages. Moreover, they address the problem of data scarcity
in offline Urdu Handwriting extraction, in the form of a data set they collected, which handles many different
handwriting styles, ligatures and writing conditions. For the data set, they use red color for handwriting as it
prevents data loss during noise removal. The methodology used is that Urdu text is passed as a scanned image,
afterwards noise is removed and the image is binarized using image processing techniques. Lastly, text lines are
segmented, and a Deep learning model extracts the information from each text line segment. The problem of
extraction is dealt as a Sequence to Sequence problem, where a sequence of letters are generated from a an input
image, which is passed as a sequence. In the paper, a one directional BiLSTM is used as it retains memory for
a longer sequence as compared to RNNs, and connectionist temporal classification loss (CTC) function is used
as the loss function. The model predicts character by character.

In \cite{anjum2020urdu_ohtr}, the authors propose an attention based approach for Urdu Handwriting extraction. In the paper,
they argue that attention based mechanisms have been successful in tasks such as machine translation, speech
recognition, and image captioning, and that attention mechanism helps extract most relevant features for images.
Therefore, in the Handwriting text extraction problem, it can help extracting the relevant features and context for
predicting the right character. In Urdu, the shape of characters change with respect to their position and shape,
hence the attention based mechanism is more relevant in that aspect. As far as methodology is concerned, the
authors deal with this as a sequence to sequence to problem, and the problem is dealt with an encoder decoder
architecture. Firstly, a convolutional neural network is used to extract features from the Urdu Handwritten
text images to get a list of vector representation of the image. Afterwards, the vectors are based into a Gated
Recurrent Unit to predict the sequence of characters in the image. In the Gated Recurrent Unit, the attention
mechanism is used, where a context vector is used to focus on the relevant part of the image. In this mechanism,
the character is predicted based upon the previous character, the hidden state and the context vector. The
context vector is also convolved with a learnable filter to smooth out the histories in the context vector. This
entire part is the decoder part, and the other part is the encoder part, where an input image is encoded to
vectors that can be fed into a GRU. To do this, the authors used DenseNet Convolutional Neural Network. In
this specific model, the convolution is applied to all previous layers, which enables diverse learning in the model.
As DenseNet model is compute intensive, bottleneck layers (which are 1 X 1 convolutions) are used inside the
DenseNet model. Lastly, CTC loss function is used in the model, alongside word beam search, which is used
to choose the best word given an input (i-e. looks for word with maximum probability). The authors collected
their own data set with the help of 100 writers to evaluate their model. They compared their model with non
attention based encoder decoder model, and their model ended up outperforming all the non attention based
models.

In \cite{UrduDeepNet}, the authors firstly collect an Urdu Handwritten text data set. In the paper, the authors explain the
3
entire data preprocessing phase for Urdu Handwriting extraction. The first phase is segmenting text by drawing
bounding boxes on the handwritten text in the images. Afterwards, the segmented text is normalized, binarized,
and denoized for text extraction. Moreover, the problem of text extraction is dealt as a classification problem,
where the input is passed through a convolutional neural network, and in the last layer, there is a Fully connected
layer which predicts characters in the image.

In \cite{ul_Sehr_Zia_2021}, the authors propose an unconstrained dataset for Urdu handwriting that covers more than 73
Urdu ligatures, NUST-UHW. Moreover, the authors also propose an encoder decoder architecture, which deals
with this problem as sequence to sequence problem. In the paper, the authors argue that Urdu Handwritten
text contains very complex features that are crucial for predicting the characters in the image, therefore low
resolution images cannot perform really well. To solve this issue, the author uses high resolution images for
Urdu Handwritten text. The images are first encoded into vectors by passing it into a Convolutional Neural
Network. In the CNN, the high resolution images are needed to be brought down to low resolution. Max pool
layers are used to achieve this. The issue with Max pool layers is that it can lead to loss of data from the image.
To address this issue, the authors concatenate the output in the last layers, instead of Max pooling. This saves
information in the last layers as the last layers in CNN hold deeper information about the image. Afterwards,
the output from CNN is passed into multiple BiLSTM layers, where again vectors are concatenated. Moreover,
an n-gram model is used to correct the language of the output from the BiLSTM layer. For the loss, CTC loss
function is used with beam search for predicting the correct word. Also, to diversify the data during training
time, a random noise layer is added before the CNN, to add random noise to the image so that at every training
step a different image is sent to the model.

\section{Methodology}
\subsection{Dataset}
In this research work, we use the Dataset collected by Punjab University College of Information technology titled as PUCIT - Offline Urdu Handwritten Urdu Lines Dataset\cite{anjum2020urdu_ohtr}.
The dataset is a multiple line written dataset written by 100 writers. In the dataset, text lines from the long written paragraphs. The dataset considers the slant nature of writing style in Urdu language. The dataset contains 7309 Urdu text lines and 78,870 Urdu words. Lastly, the dataset is available for only non-commercial academic purpose
\subsection{Preprocessing}
The images in the dataset were not normalized, there were images with different sizes and channels. We standardised it by applying image processing techniques. Moreover, we also applied some image binarization techniques for better results.
\begin{figure}[h!]
  \includegraphics[width=\linewidth]{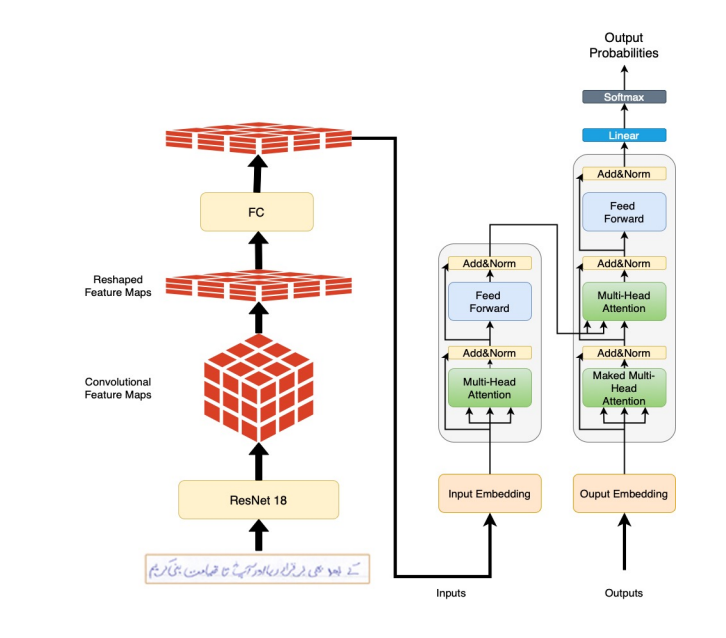}
  \caption{Resnet-Transformer Model Architecture for Urdu Handwritting Extraction}
\end{figure}
\subsection{Procedure}
We addressed this problem as a sequence to sequence problem, just like in the previous studies, but instead
of using LSTMS or RNN, we used Transformer architecture \cite{DBLP:journals/corr/VaswaniSPUJGKP17} for predicting the text in the image.
Transformer architecture has been popular in the NLP domain as it was able to provide state of the art results in
NLP tasks, such as Machine Translation, Question Answering, Named Entity Recognition, Language modeling
etc. Transformers are quite good in understanding inter language dependencies, and it also learns a language
model for output text. In the transformer architecture, there are two blocks, encoder and decoder. In the encoder
part, there’s a self attention mechanism, which focuses on different features of the input for every segment of the
input image, whereas in the decoder part there is causal attention mechanism, that only looks into the previous
tokens. There also exists attention mechanism between the encoder and decoder. These attention mechanisms
can help handwriting problem by focusing on important part of the image for predicting the characters, as
Urdu characters are context dependent. Moreover, the language modeling capability of Transformers can help
in correct language prediction in the same pass as text extraction. There’s no need of separate language model.
Moreover, RNNS are not parallelizable, but transformers are parallelizable, which can make training large models
scalable. Before passing an input image to a transformer, it needs to be converted into a sequence of vectors
as it is required by the transformer architecture, so we used ResNet\cite{DBLP:journals/corr/HeZRS15} 18 architecture for extracting
high dimensional features from the input image. As Urdu is a complex language, there is a need for extracting
complex features, for which ResNet does a good job, as in ResNet there are residual blocks, which retain the
information of previous layers as we go deep into the network, so information is not lost. The problem is
dealt as a classification problem, where a sequence of characters will be predicted from the Transformer layer.

In this approach, we first take an image, and pass it through a Resnet18 model to extract important features from the image. Afterwards, we pass the output of the Resnet18 model into Linear layers (we tried different combinations) to get a vectorr (embeddings) representation of the images. We then passed these representations into the transformer model. The transformer model predicts character by character. We trained the model with masking the target. Resnet18 model was pretrained on imagenet, which was finetuned during the training process, while on the other hand transformer was built from scratch without any pretrained weights.
\begin{figure}[h!]
  \includegraphics[width=\linewidth]{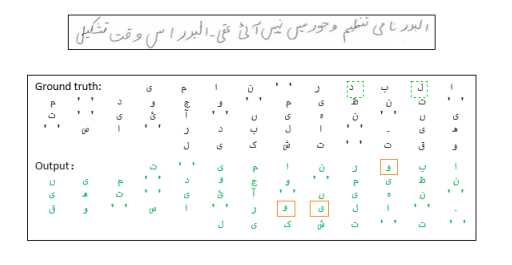}
  \caption{Sample Input and ground truth for training of dataset by \cite{anjum2020urdu_ohtr} }
\end{figure}

\section{Evaluation and Experiments}
For the evaluation, we used character error rate (CER) between input and output text. It is illustrated below.
\begin{equation} \label{eqn}
	CER = \frac{ins + sub + del}{n} \times 100  
	\end{equation}
It is the number of insertions, substitutions, and deletions required to transform output text into the target
label. While n is the number of characters in the target label. This equation can also be used to calculate word
error rate. The model will be considered good if it achieves low CER or WER.

The model was ran on a machine with the following specifications:
\\\noindent\\\textbf{Processor}: AMD Ryzen 5 3600 @4.2 GHz  6 Cores/12 Threads\\\textbf{RAM}: 32GB DDR4 3200 MHz\\\textbf{GPU}: Nvidia GeForce RTX 3070 Ti 8GB\\

Despite transformers showing great results in Natural Language Understanding, there exists one limitation, that it needs huge amounts of data to train. Unfortunately, that is the limitation of our research, as there is a substantial lack of availability of large Urdu handwritten text datasets. We trained our model for around 12 hours, however, it didn't converge. Due to very limited data, we achieved more than 85\% error rate. In our research, we were able to explore transformer architecture for handwritten text extraction, but due to a lack of large datasets, we were not able to achieve good CER. Moreover, we trained the transformer on character level, due to which output sequence had a huge length, which may be the reason for not fitting well on the data. In the future, we can test on pretrained models, where we can use a pretrained Urdu Language decoder, and train our image feature extractor for better results. Lastly, our research has opened pathways for researchers to work on transformer based OCR for low resource languages, such as Urdu.

\acks

Firstly, we would like to thank Almighty Allah (S.W.T) for granting us the strength to work on this project. Moreover, we would like to thank the contributors to the Urdu datasets that we employed the use of. Lastly, we would like thank our instructor, Dr. Mirza Omer Beg for giving us the opportunity to work on this problem, while making us stay motivated throughout the entire semester.

\nocite{*}

\bibliography{aim}

\bibliographystyle{abbrvnat}

\end{document}